# Midgar: Detection of people through computer vision in the Internet of Things scenarios to improve the security in Smart Cities, Smart Towns, and Smart Homes

# Midgar: Detection of people through computer vision in the Internet of Things scenarios to improve the security in Smart Cities, Smart Towns, and Smart Homes

Cristian González García[a*], Daniel Meana-Llorián[a], B. Cristina Pelayo G-Bustelo[a], Juan Manuel Cueva Lovelle[a], Nestor Garcia-Fernandez[a]

[a] *University of Oviedo, Department of Computer Science, Sciences Building, C/Calvo Sotelo s/n 33007, Oviedo, Asturias, Spain. Tel: +34985103397*

[a] gonzalezgarciacristian@hotmail.com, *danielmeanallorian@gmail.com, crispelayo@uniovi.es, cueva@uniovi.es, nestor@uniovi.es*

* Corresponding author.

*Abstract*— Could we use Computer Vision in the Internet of Things for using pictures as sensors? This is the principal hypothesis that we want to resolve. Currently, in order to create safety areas, cities, or homes, people use IP cameras. Nevertheless, this system needs people who watch the camera images, watch the recording after something occurred, or watch when the camera notifies them of any movement. These are the disadvantages. Furthermore, there are many Smart Cities and Smart Homes around the world. This is why we thought of using the idea of the Internet of Things to add a way of automating the use of IP cameras. In our case, we propose the analysis of pictures through Computer Vision to detect people in the analysed pictures. With this analysis, we are able to obtain if these pictures contain people and handle the pictures as if they were sensors with two possible states. Notwithstanding, Computer Vision is a very complicated field. This is why we needed a second hypothesis: Could we work with Computer Vision in the Internet of Things with a good accuracy to automate or semi-automate this kind of events? The demonstration of these hypotheses required a testing over our Computer Vision module to check the possibilities that we have to use this module in a possible real environment with a good accuracy. Our proposal, as a possible solution, is the analysis of entire sequence instead of isolated pictures for using pictures as sensors in the Internet of Things.

***Smart Cities; Smart Towns; Smart Homes; Internet of Things; Smart Objects; Computer Vision; Surveillance; Security;***

## I. INTRODUCTION

Currently, we live in the information era. We have many things in our daily life with access to the Internet and capable of making our daily life more comfortable like smartphones, tablets, computers, some cars, Smart TVs, and so on. Every new day, we have more devices and better Internet connection [1]. These devices are able to run programmes, which use the devices' sensors, or to do other tasks like creating alarms or notifications, turning on or turning off the device, and et cetera. These objects are known as Smart Objects [2].

Smart Objects provide us with many possibilities and every day we have more different Smart Objects. However, Smart Objects can be more useful in the case of being interconnected with each other and with other objects like sensors or actuators [2]. This interconnection is called the Internet of Things (IoT). The Internet of Things allows creating huge or small networks in order to obtain a collective intelligence through the processing of objects' information. The first example is how the IoT was born because the first idea was its implementation in supply chains [3]. Other examples are the object identification in chemistry plants, people, and animal using Smart Tags as Radio Frequency IDentification (RFID) and Near Field Communication (NFC) [4–6]. The IoT can be applied in cities, also known as Smart Cities, in order to offer different services that improve citizens' 'livability' [7]. Smart Towns use a similar application although they use the IoT to preserve their culture, folks, and heritage in small cities and towns [8]. In other cases, we can use the IoT to create Smart Homes that can control and automate certain things of our houses like doors, windows, fridges [9–11], irrigation systems, lights, distribution multimedia [12], and so on. By the contrary, some governments apply the IoT to control and care in the better way the Earth, which is also known as Smart Earth [13], in front of different dangers like fire, earthquakes, tsunamis, or floods. Moreover, we can use the IoT to anticipate and prevent human disasters like the case of the Deepwater Horizon at Gulf of Mexico [14] or the problem in the security system of the Nuclear Central of Fukushima to detect the tsunami and automate the turn off of the diesel motors or activate an protection over the motors. Nonetheless, these systems need a central service to control and manage their data and their objects. Besides, sometimes these systems need to create intelligence and take the decisions. Thus, they need an IoT platform.

There are several IoT platforms for many and different uses: business platforms, research platforms, platforms in beta state, and open source platforms [15]. All these platforms are more or less similar because they allow working with objects and interconnecting them. Some platforms offer people an Application Programming Interface (API) to facilitate this task.

A very interesting application for the IoT would be the recognition of people using Computer Vision. Integrating Computer Vision in an IoT platform could offer more security systems to Smart Cities, Smart Towns, Smart Homes, and Smart Earth because they could recognise a person in the incorrect place at the wrong hour because, maybe, he is a thief or a potentially dangerous person for the environment like a pyromaniac. We could obtain this functionality without the need of a person to supervise the camera and facilitate this labour to revise only the critical pictures. We can see a similar research in [16], where they use sensors to obtain data and in the case of these data accomplish certain conditions then the camera will take pictures. A similar idea was proposed in [8] in order to protect the heritage of Smart Towns. Notwithstanding, they did not apply Computer Vision because they only take a picture under certain conditions. After that, they need a person to see that picture and evaluate the situation.

For these reasons our hypotheses were: Could we integrate Computer Vision in the Internet of Things? Could we use the pictures from an Internet Protocol (IP) Camera as a sensor? Could we obtain a good accuracy to automate or semi-automate this kind of events?

A possible solution for solving these hypotheses is the creation of a Computer Vision module and the integration of this module in an IoT platform. In our case, we used the Midgar IoT platform [15,17]. We developed a Computer Vision module and modified Midgar to support Computer Vision. To test our first one hypothesis, we used an IP camera connected to the IoT platform and we took different pictures to test the functionality. To test the last both hypotheses, we took many pictures of the inside of the laboratory. After, we tested all these pictures in a batch process with our Computer Vision module to obtain the accuracy that the module has. However, in order to improve the possible detection of people, we analysed the entire sequence to find people instead of analysing picture by picture in the way of improving the identification of people by the use of all pictures of a movement.

In the rest of this article, we will discuss, in section II, about what the Internet of Things and Smart Objects are, and explain the different IoT applications like Smart Cities, Smart Towns, and Smart Homes. Besides, we will explain a brief of Computer Vision, we will continue talking about the current and more relevant IoT platforms, and we will present the related work. In section III, we will describe the case study, in our case, the Midgar IoT platform. After, in section IV, we will explain the methodology that we used and show all the results of our evaluation and the discussion of the results. To finalise this paper, we will present our conclusions in section V and show possible future work to do from this research in section VI.

## II. STATE OF THE ART

Nowadays, the Internet of Things is one of the most used technologies with interest for some countries, like the United States of America [18] and the United Nations [5]. Nevertheless, the IoT needs many improvements because it was born a few years ago and has many problems to resolve as we can see in different recent articles about Smart Towns [8], protocols [19], security [20], or others [5,12]. The goal is to use the IoT to interconnect everything, from food to computers and then, automate different processes to improve our daily life. Now, we have different Smart Objects with a small but smart functionality in our life. Besides, we can use these devices to make our life easier in some Smart Cities with special services like Santander [21], for instance, to park or manage our Smart Homes, or automate some tasks like the irrigation service. However, sometimes this is very difficult. For example, if we have a burglar alarm in our house and we have pets, we would have a problem because, maybe, the alarm would sound due to the animal whereas we only want that the alarm sound when the alarm recognises a person. We propose a solution based on the use of Computer Vision through the Midgar IoT Platform [15,17].

### A. Internet of Things

The Internet of Things allows the interconnection of physical and virtual things. These things can be objects of the physical world or information of the virtual world. This interconnection can be at any time, maybe meanwhile you are moving around the world, in continuous motion, or at any time during the day, in any place like outdoor or indoor, and between anything like Human to Human (H2H), Machine to Machine (M2M), or between Humans and Machines (H2M) [22]. The IoT allows creating a Smart World, which is the fusion of heterogeneous and ubiquitous objects, Smart Cities, Smart Towns [8], and Smart Homes [23], with all devices capable of interacting between themselves [8]. The IoT has originated the development of object automation through the Internet to exchange information [24], allowing the creation of this Smart World.

However, many heterogeneous things compose the IoT. The most important are the Wireless Sensor Networks (WSN), which are the core of the Internet of Things [25]. A WSN interconnects sensors, in order to obtain data, with a server or special system to work and maybe, automate tasks in one place. Other components are actuators, which allow executing actions, like motors, fans, machines, and so on. Another type of network is the fusion between a WSN and the actuators, known as Wireless Sensor and Actuator Network (WSAN). Besides, Smart Objects are other important components because they can perform actions as actuators, they can sense because usually they have sensors, and they are smart to process information or data and perform actions. Nevertheless, all these components need a connection with the Internet, but currently, almost every objects can be connected to the Internet [26].

This is why the definition of the IoT is the next: The Internet of Things is the interconnection of heterogeneous and ubiquitous objects between themselves through the Internet [5,6,15,17]. The goal of the IoT is to interconnect the whole world through the creation of different smart places to automate, improve, and facilitate our daily life [17].

*B. Smart Objects*

A Smart Object or Intelligent Product is a physical object with the capacity of interacting with the environment, in some cases with intelligence to make decisions, with autonomous behaviour, and identifiable in its whole useful life [2]. As we said before, Smart Objects are one of the fundamental parts of the IoT. Some examples of Smart Objects [2] are smartphones [27], Smart TVs [27], tablets, some cars [27–29], IP cameras, computers, microcontrollers, some freezer prototypes, and etcetera.

However, some objects, which are connected, have a limited memory, CPU, or power [26]. For instance, an IP camera might not have the necessary space or power computation to run applications with the capabilities of applying Computer Vision to one or more pictures. Thus, in these cases is needed other resources to obtain the necessary computing. One option is to use an IoT server capable of offering that intelligence through the network as occurs with the IoT platform Midgar [17]. In this way, we can expand the possibilities of objects like Smart Objects or Not-Smart Objects [2].

We can classify the Smart Objects according to the three dimensions according to [2]: **Level of Intelligence**, **Location of the Intelligence**, and **Aggregation Level of Intelligence**. In our case, our IP camera is a Smart Object with a level of intelligence of **Notification of the Problem**, with the **Combined Intelligence**, and **Intelligence in the Item**.

*C. Smart Cities, Smart Towns, and Smart Homes*

Smart Cities are a very important part in the IoT [30]. A Smart City has different types of sensors distributed around the city to gather information about the city. Through the use of ubiquitous communication networks, WSN, WSAN, and intelligent systems, Smart Cities can offer new services to citizens [31], and facilitate their daily life and improve the city 'livability' [32]. We can see some European Smart Cities like Luxemburg, Aberdeen, Oviedo, and many other cities with their qualification in based on different criterions in [33] or in [34], which define the Smart City concept and analyse the Smart Cities of Europe in based on six indicators. Another example is a big European project called SmartSantander [21], which proposes an IoT architecture for Smart Cities and shows the different services that this architecture can offer.

Moreover, Smart Towns also exist. They are small cities or towns with a great culture and heritage that need to preserve and revitalise instead of only improving the 'livability' as Smart Cities. Smart Towns have to protect and expand their culture and heritage to avoid the oblivion of their buildings, monuments, landscapes, folklore, tradition, and a long etcetera. For instance, Smart Towns can be capable of sharing their places, recording their culture and the way of making their typical dishes, monitoring the conditions in a specific place that needs special conditions like libraries or museums, or protecting the monuments [8].

Smart Homes [35], also known as Intelligent Homes, are closer to be a reality searching the 'livability' in our homes. Smart Homes provide us with an automated system to improve our daily life at our homes. They are based on a WSAN that allows controlling different objects at home performing certain events not only in an automatically way but also invoked by remote controls or smartphones. We can control the doors or the windows using our smartphones or Smart Tags like RFID or NFC, create an automate system lights, save money in the heater with different sensors and an intelligent system [36], and so on.

Through this proposal, we want to improve the $4^{th}$ and $5^{th}$ principles of the 'livability' [7,8]. We want to improve the security in these three IoT areas using Computer Vision for detecting actions, people, or certain behaviours. By means of using an IP camera, we are able to send pictures when we detect moving objects, removed objects, abandoned objects, and manipulations of the camera. After, we can apply Computer Vision to these pictures in order to detect dangerous situation for important things like our families or things in homes, monuments, heritage, specific people in towns, and some type of citizens in cities.

*D. Computer Vision*

Computers can only process zeros and ones. Nevertheless, years ago, the Artificial Intelligent (AI) was born to offer the possibility of creating programmes that allow computers to learn. John McCarthy coined this term in 1955 in the conference of Dartmouth [37]. Inside the AI, one of the fields is the Computer Vision. Computer Vision is the field that allows computers to 'learn' to recognise a picture or the characteristics of a picture. This allows identifying objects, humans, animals, or a position in a picture. Thus, the goal of the Computer Vision is that a machine can understand the world [38].

For reaching this goal, there are many algorithms for the long process of recognising something in a picture. Some algorithms to obtain the features of the dataset that you can use to train the model are Histogram of Oriented Gradients (HOG) [39], Local Binary Patterns (LBP) [40–42], HOG-LBP [43], Scale-Invariant Feature Transform (SIFT) [44], and Speeded-Up Robust Features (SURF) [45]. Other algorithms are the necessaries to train the model using the extracted features obtained previously. Some examples of these algorithms are Support Vector Machine (SVM) [46] and Logistic Regression [47,48]. However, the task of obtaining a good model is very difficult. You need to take many good pictures and try many times with another group of different pictures to check that your model works well. Besides, you need to create a model to solve your problem, because the use of general models could reduce the accuracy. Examples of these models could be the sample models of some applications like OpenCV. Even though, in this paper, we will use this type of models.

In our proposal, we use Computer Vision to recognise the presence of people in the pictures that the camera sends to the IoT platform in order to perform an action in the positive cases. Thus, we want to use the pictures as a special sensor.

*E. Internet of Things Platforms*

As we explained before, to obtain the best potential of Smart Objects, we need to interconnect them between themselves. Notwithstanding, we need a 'brain' which can manage and notify the Smart Objects and sometimes, to work as the brain for some other objects like actuators. This 'brain' is an IoT platform. For this purpose, there are different IoT platforms with different pros and cons. We can classify these IoT platforms in the next four groups [15]:
- Business platforms: Xively [49], Exosite [50], SensorCloud [51], Etherios [52], ThingWorx [53], Carriots [54], Azure IoT Suit [55], Amazon Web Services [56], and IBM *Internet of Things* [57].
- Research platforms: Midgar [17], Paraimpu [58], QuadraSpace [59], SenseWeb [60,61], and SIoT [62].
- Platforms in beta state: Sensorpedia [63,64], Evrythng [65], and Open.Sen.se [66].
- Open Source platforms: ThingSpeak [67], Nimbits [68], and Kaa [69].

Some of these IoT platforms have characteristics that others do not have. However, none of them has a module of Computer Vision that allows working with pictures as sensors. You can use an IP camera as an actuator, namely, you can connect the IP camera and take pictures under certain conditions. Our intention is to use the IP cameras' pictures as sensors. For example, you could connect the IP camera and send the pictures when a certain condition was accomplished. Then, when the IoT platform received the pictures, the IoT platform would have to analyse the picture for searching, for example, people. In the case that the IoT platform detected people, the IoT platform would trigger the action that the user had defined for this case. For that, our proposal is one possible solution to use the IP camera as a sensor using Computer Vision to detect a specific thing in pictures.

*F. Related Work*

In the current literature, there are some uses of cameras in combination with IoT and sensors. In some cases, they use this combination to improve the job conditions, obtain more data without travelling to the place, or to obtain knowledge about something.

One of those uses is to improve the care of bees and facilitate the job of beekeepers [16]. They used a sound sensor to send a picture when the sound exceeded some limit, and in this case, it would send a message with that picture to the beekeeper. Then, the beekeeper could decide if the hive needs his visit or not depending on the things that he saw in the picture and the sensor information obtained through a WSN. With this system, the keepers could reduce the frequency of their visits to the moments that they receive critical information, as demonstrated in [16] because they can see information remotely. They analyse the sensor information with an algorithm to avoid the human interaction for obtaining the state of the beehive but, just the same, they have to see the picture when they receive it to see what happen in the hive.

Another example is the proposal of using this combination for learning. For instance, the IoT could help to learn and show different knowledge between master and students by collecting data and find the best way to train. With this way, they can help to protect the heritage and folks of towns [8].

Other examples of Computer Vision is when it is used to create maps or study maps, which is called Cartography [70]. A clearly Cartography example are Google Maps or Bing Maps, which modify the maps to give a service to people. In this way, Computer Vision is applied to recognise some specific parts in maps, like roads, buildings, water, or fields. This is an example of how to applied Computer Vision in Smart Earth.

Another possibility is to use the Computer Vision as these authors show in [71]. In this article, the authors propose the use of Computer Vision to simulate the sight of humans combined it with other sensors to simulate the five senses of the human body. Exactly, their idea is to combine a camera, which allows identifying things, with different sensors. This combination could calculate the distance to the objects that the camera can see and interact in the Internet of Things.

The previous proposals used the IoT with cameras. However, they needed a person to see the picture in order to make a decision in the first case or they record the movement to add more information to the sensors in the second one case. In our proposal, we use the camera to obtain the picture and then, we send the picture to a Vision Computer module. This module is the responsible for taking the decision and sending this decision to the IoT network, which is the manager of the service. Then, we automate this step according to a model in order to avoid the intermediary and accelerate the response because maybe, in some cases, it is impossible for people to see many pictures or take a decision immediately.

III. CASE STUDY: MIDGAR

Midgar is an Internet of Things platform to investigate different solutions for the IoT [15,17]. In this paper, we try to find a solution for the integration of Computer Vision in an IoT platform for analysing pictures from IP cameras in order to find a determinate object in the pictures and use the pictures of the IP camera as sensors. In this section, we are going to describe the changes that we did in the Midgar platform and show our proposed solution to add the Computer Vision module in Midgar.

*A. Midgar Architecture*

The system architecture is very similar to the original Midgar architecture. It has the same four layers as we can see in Figure 1: **Process Definition**, **Service Generation**, **Data Processor and Object Manager**, and **Objects**. However, we added the Computer Vision module in the third layer and we had to modify the different layers to support the new functionality.

The first layer is the **Process Definition**, which contains the user's process. This is the only layer with user interaction. The user (Figure 1.1) must define the process that he needs through **Midgar Object Interconnection Specific Language** (**MOISL**) that was developed in [17] and we can see in Figure 1.2. **MOISL** was developed using the HTML5 canvas. When the user finished the definition, and click the generate button, the editor generates the **Serialised Model** (Figure 1.3). Then, the editor serialises the model that the user has defined using **MOISL** in an eXtensible Markup Language (XML) file. This **Serialised Model** contains all the necessary information about the model, which was created by the user, and the information that the second layer needs to generate the **Active Process**.

Afterwards, the **Service Generation** receives the **Serialised Model**. This second layer parses and processes the information of the **Serialised Model** in the **Processor** (Figure 1.4), which creates, compiles, and executes the **Active Process** that interconnects the objects (Figure 1.5).

The **Active Process** is placed in the third layer, **Data Processor and Object Manager**. The **Active Process** keeps working in the server while is performing the defined user's task. This process has a continued and direct communication with the **Midgar Store** (Figure 1.6), which is a part of the Midgar core because the **Midgar Store** contains the database with the services, the objects, the actions, and the data.

The last layer is the layer that contains the **Objects**. In this case, our IP camera. These objects implement the message interface to keep a permanent and bidirectional connection with the server (Figure 1.7). However, the IP camera cannot implement this message service because the IP camera software is private. Nevertheless, it can send pictures by HTTP protocol. Then, the IP camera has to send the picture using the REpresentational State Transfer (REST) of Midgar service. After, the Midgar service, which is in the third layer, realises that this is a picture since Midgar analyses the Multipurpose Internet Mail Extensions (MIME) type of the request and sends this request to the **Computer Vision module** (Figure 1.7). The **Computer Vision module** analyses the picture and responds if the picture has or not a person.

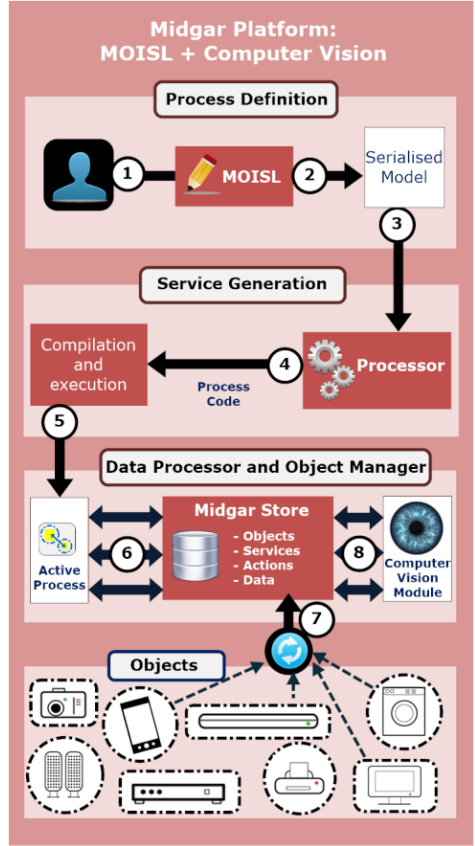

**Figure 1 Midgar Platform architecture with the Computer Vision module**

*B. Implementation*

In this subsection, we are going to explain the new implementation of Midgar. Firstly, we are going to explain the flow through the different layers. Afterwards, we are going to describe the functionality and the interconnection of our Computer Vision module in Midgar. Lastly, we are going to talk about the functionality that the IP camera offers.

*1) Midgar flow*

In order to add the capacity of Computer Vision to Midgar, we had to create a module with this capacity. The rest of Midgar is equal to the previous platform [17]. Then, the difference is when Midgar receives a picture. When this happens, Midgar detects that the request contains a picture because Midgar is able to analyse the MIME type of the request. For instance, when an object like an Arduino or other, which has the possibility to deploy applications, is connected to Midgar, this object can send a message using the XML standard style of Midgar. In this case, the Canon IP camera does not allow modifying the software, as occur with others IP cameras, but we can analyse the MIME type to see if it is a picture as we can see in Figure 2. Then, when Midgar receives a picture, Midgar saves the picture in a folder. When Midgar spends five seconds waiting for pictures and more pictures have arrived, Midgar sends the picture sequence to the Computer Vision module. After that, the Computer Vision module analyses the folder, which contains the whole picture sequence, to find people in at least, one picture. In the case that the module finds one or more people, it will respond to Midgar with a 'True', in another case, with a 'False'. Then, Midgar will store this response in the database, as if it were a sensor with only these two possible states.

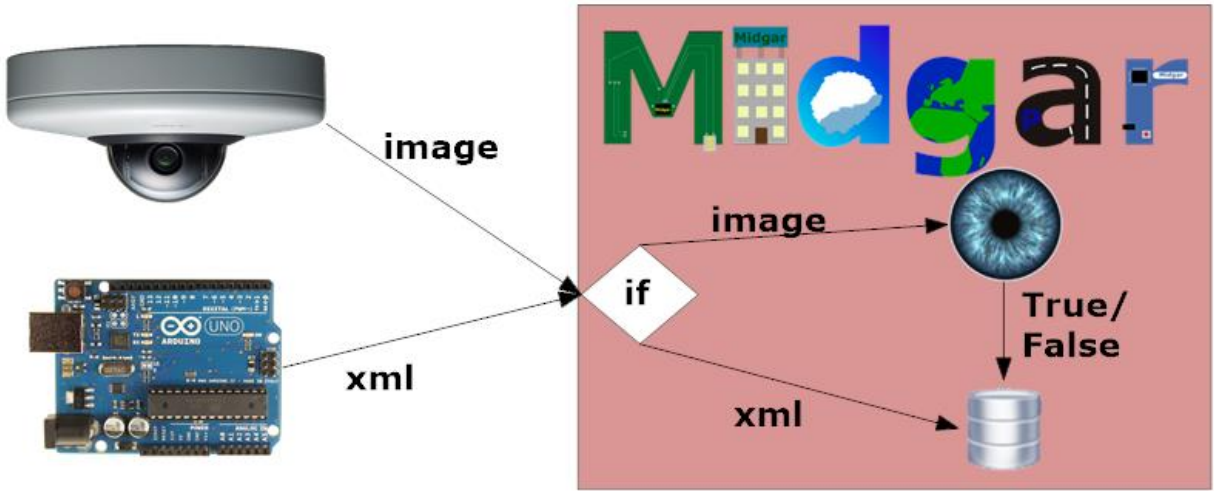

**Figure 2 Midgar flow**

*2) Computer Vision Module*

We chose as a possible solution the creation of a separate module. In this way, we could call this module when we need to evaluate a picture or a picture sequence. The decision was to separate the implementation of the Computer Vision module from the IoT platform. This allows us to execute different tests with the same module and the same architecture for the evaluation of this proposal and then, we only have to change the parameters that we use to call the Computer Vision module.

The Computer Vision module was developed in Python due to the requirements of the Open CV library that performs the body detection work. The use of Open CV also required the use of the library Numpy. The workflow of this module consists of loading an image from a file, converting the image to a bytes array, transforming the image to grey scale, and using the OpenCV library to detect the number of bodies in the image. If there is anybody, the module will return the Boolean value 'True' else, the module will return the Boolean value 'False'. However, we chose to improve the module recognition by using of picture sequences instead only one picture. In this way, we could obtain more accuracy because our objective is the detection of a dangerous movement.

However, OpenCV needs to setup a few variables. The first one is to indicate the scale factor. The scale factor is necessary to create the scale pyramid that the algorithm uses to find objects in different depth inside the picture, which we set this value to '1.01'. The second variable is the minimum near detections that are required to compose a single object. We set up this second one with the value '10'. The last parameter is the minimum size of each detection window that we set up to '(200, 200)'. Nevertheless, the values that we used to set up the OpenCV library depend on the context. Furthermore, an external XML file is required because OpenCV loads the classifier from an external file in order to reuse the same code with different classifiers. In our proposal, we decided to use three sample classifiers that allow OpenCV to detect upper bodies, frontal faces, and the combination of heads and shoulders. If we wanted to detect other things, we would create new classifiers extracting the needed features. Furthermore, these models are examples and they are a bit weak. For this reason, if we want a better recognition, we should train a new model to obtain a better and more specific classifier. However, we could improve in any way the movement detection in the case that we analyse the sequence to obtain at least one picture, which means that it is a positive detection. For instance, in the case that we would have a better classifier, we could increase this number and require at least three or five pictures with the object that we want to recognise. It is very useful if, for instance, we want to use this system to detect dangerous people like burglars, thieves, and so on in our home, or people in some private or dangerous area. In these cases, it is preferred that the module gives false negatives instead false positives cases. The reason is because if we received a false negative, we only received a false alarm. On the contrary, if we obtain a false negative, we could have a thief in our home.

In order to test the body detection, another module was developed that skips Midgar. This module, also developed in Python, uses the library Flask to receive the images from the camera and follows the workflow of the other module although it does not return a Boolean value. This module saves the images that the camera sends in a directory and saves another image if the module detects a body, drawing a green rectangle around the body in the image. With this information, we will be able to do the evaluation of our proposal.

*3) Canon IP Camera*

We used the Canon VB-S30D as IP camera but before, we had updated the firmware to the last version, the 1.2 of May 26[th] of 2015. We connected the camera through Ethernet connection. This camera is a Smart Object because it recognises its own data and it can make decisions according to the video. This IP camera allows doing streaming using a URL or send pictures or

emails when the IP camera detects some changes in the video. For instance, if it recognises some movement or the modification of some object in the scene, and it can send pictures or emails with the pictures of that moment.

In Figure 3 we can see the five detection types that the camera offers. The first one is the 'moving object detection' and this consists of detecting some movement. Secondly, the 'abandoned object detection' which consists in notifying the presence of new objects. Another is the 'removed object detection' which detects when any object disappears from the default scene. The fourth is the 'camera tampering detection' which consists in detecting when the camera was manipulated. The last one is the 'step detection' which detects a movement over a defined line. These five events are configurable and allow us to receive pictures or emails only when the scene was changed.

Besides, we can choose two modes in the camera. We can analyse the streaming or analyse the pictures that we received. For working with the camera in Midgar, firstly, we had to modify the IP camera configuration by setting the IP and port of the Midgar REST service, where the camera had to send the picture. Afterwards, we had to register the IP camera in the platform so that the camera could be selected in MOISL. Then, we could select the camera in MOISL, create the interconnection, and work with the camera.

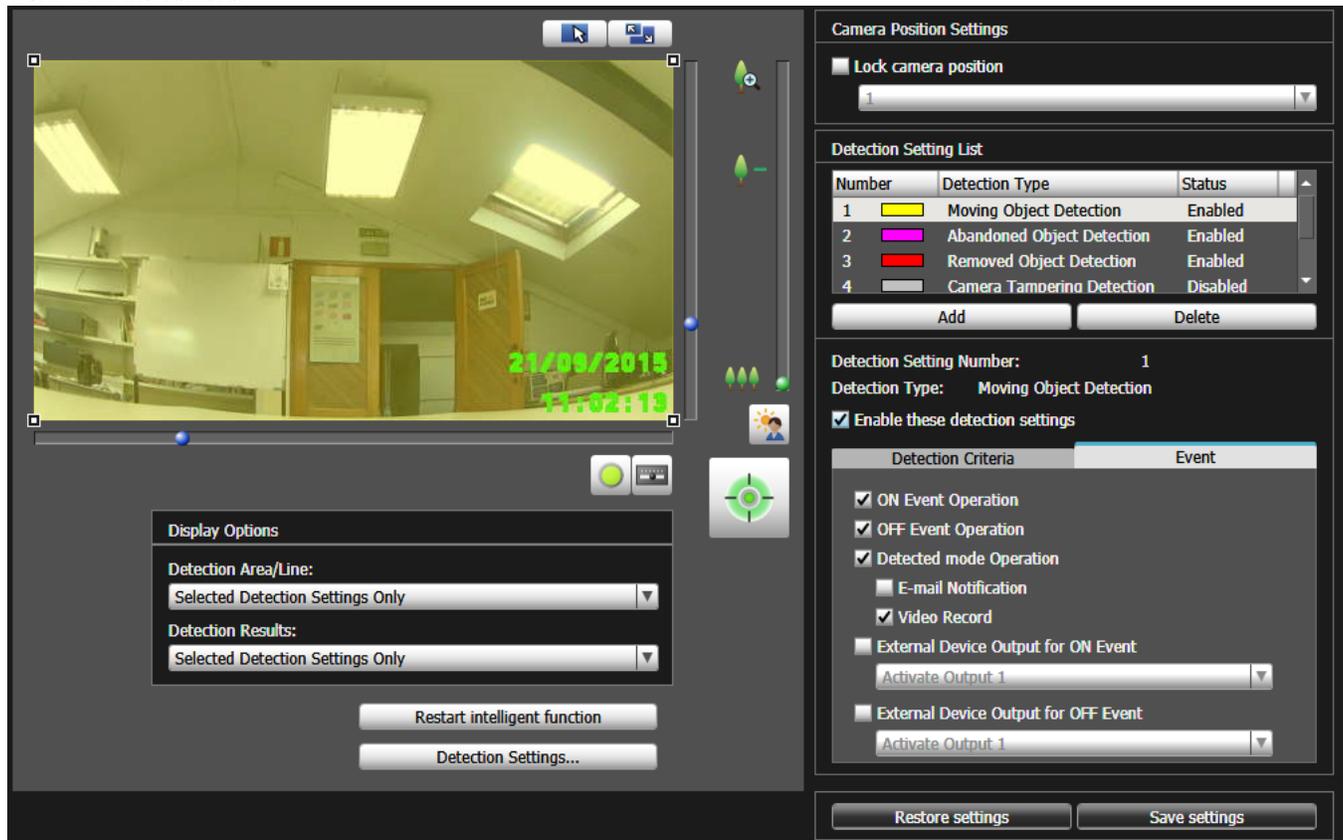

Figure 3 Canon IP camera with the 'Moving Object Detection' selected mode

*C. Used Software and Hardware*

We used the next software to develop this research work:
- Midgar:
    o The Midgar server is based on Ruby 2.3.1p112 and it uses the Rails framework 4.2.6.
    o Thin web server 1.6.4
    o MySQL Database 5.5.53
    o The graphic DSL, MOISL, was developed by using the element HTML5 canvas and JavaScript.
    o The application generator module was developed using Java 8.
- Computer Vision module:
    o The computer vision module was developed using Python 3.5.2
    o The library OpenCV 3.1.0 to apply Computer Vision to the pictures
    o Numpy 1.11.3 because it is a requirement of OpenCV
    o Flask 0.11.1 to develop a mini-server to the test

- Models:
    - Frontal face: 'haarcascade_frontalface_alt_tree.xml' [72]
    - Head and shoulders: 'haarcascade_head_and_shoulders.xml' [73]
    - Human body, Pedestrian Detection (22x18 upper body) 'haarcascade_upperbody.xml' [72]
- Camera IP:
    - Canon VB-S30D with firmware v.1.2.0
    - Internet Explorer 11 to access to the camera configuration

For the evaluation of the proposal, we used the next hardware components:
- One Raspberry Pi 2 Model B as a dedicated server with Raspbian 4.4.38-v7+
- Three Android smartphones: A Nexus 4 running version 5.1.1, a Motorola with version 2.2.2, and a Samsung Galaxy Mini S5570 with version 2.3.6.
- One Arduino Uno microcontroller board based on the ATmega328.
- During the various tests, we used as actuators: a speaker, a servo-motor, a DC motor, and several LEDs.

## IV. EVALUATION AND DISCUSSION

In this section, we describe with all detail the methodology that we did to evaluate our hypotheses. After that, we show the results that we obtained in our evaluation. We have divided the evaluation into two phases: manual pictures and automatic pictures. Moreover, we have compared the results of three different models in order to conclude what model was better for our proposal. We are going to explain each subsection through these two phases.

### A. Methodology

The main objective of this evaluation process is to verify our hypotheses:
- Could we insert the use of Computer Vision in the Internet of Things?
- Could we use the pictures from an IP camera as a sensor?
- Could we obtain a good accuracy to automate or semi-automate this kind of events?

We have demonstrated the possibility of the use of Computer Vision in the Internet of Things in the previous section, in the Implementation. Then, we have demonstrated our first hypothesis. Now, we are going to explain how to try to validate if our second hypothesis is possible. For validating the second hypothesis, we used two different phases but both with the same objective: to evaluate the accuracy of the Computer Vision module using pictures to detect people. In both phases, we used pictures without people and pictures with people and three different models. Then, we used the pictures without people to detect the false positives and true negatives cases for each model. With the pictures with people, we also obtained the false negatives and true positives cases for each model. For these both phases, we used the module without the Midgar interaction because we need to obtain the picture with the green rectangle to evaluate in a quantitative way if our module works well. Our two phases are the following:
- **Phase 1 - Manual pictures**: in this first one phase, we used pictures that we took with the Canon IP camera inside our laboratory manually. In this phase, we tested the Computer Vision module with isolated pictures without any relationship with the rest of pictures.
- **Phase 2 - Automatic pictures**: in the second phase, we used the picture sequences that the camera sent us when the camera detected some movement in the laboratory with its sensors. In this case, the pictures of the sequence to evaluate have a relation between themselves because the pictures belong to the same movement. With this case, we wanted to try the Computer Vision module with a picture sequence to improve the detection algorithm using the relationship of the pictures of one same moment.

For these both phases, we used the three different models in order to compare the different results and conclude with is the best model to use to our proposal. These models can detect the upper body of people, the frontal faces, and the combination of heads and shoulders. They have low accuracy because they are general models and need pictures with people in the correct position. If we needed a better accuracy, we should create our own specialised model. For taking the pictures that we used in this evaluation, we placed the IP camera in the middle of our laboratory in a position to watch the entrance door as we show in Figure 4.

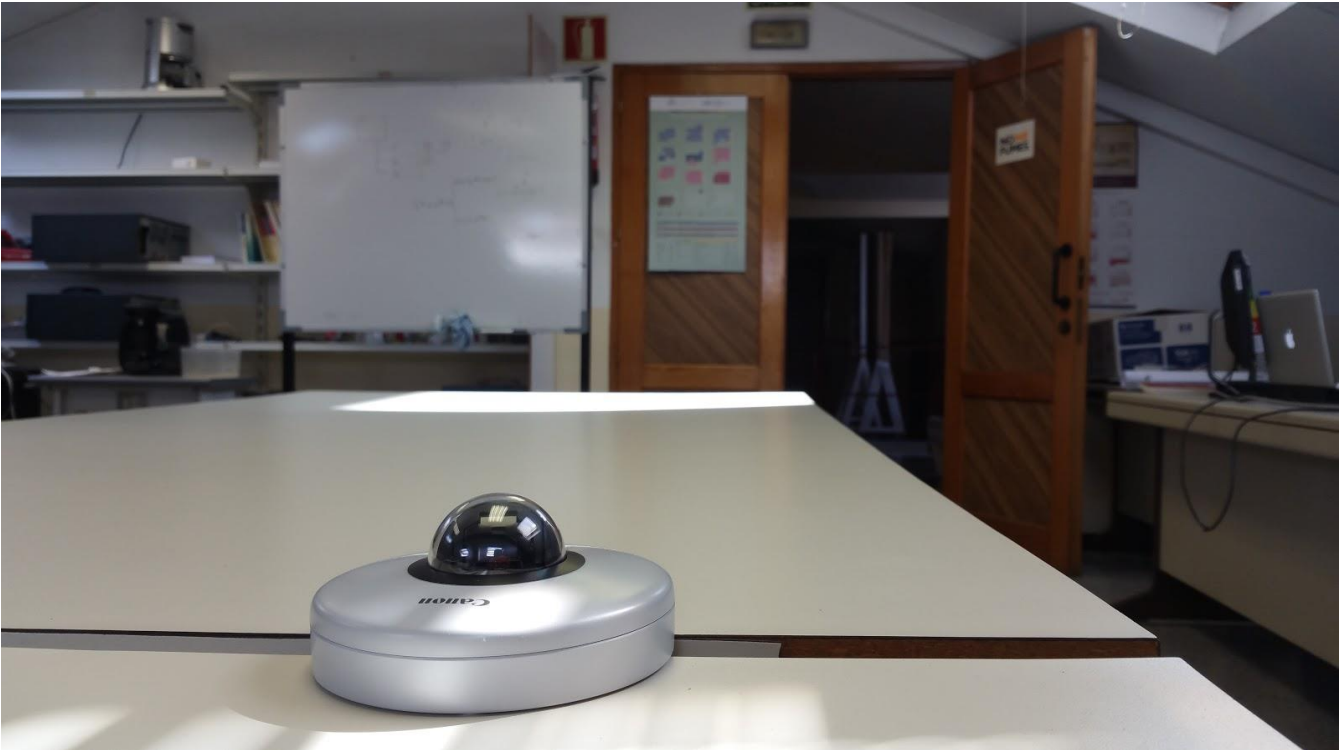
**Figure 4 Canon camera IP situation in the laboratory with the background that we used to take the pictures**

*1) Phase 1: Manual pictures*

For this first phase, we used the manual mode of the IP camera. We divided the pictures into two folders according to pictures without people and pictures with people. We analysed each folder with our module. For each folder, we obtained a new folder with the detected person inside a green rectangle. After it, we reviewed manually each picture taking into consideration the expected result because, maybe, the green rectangle could mark an incorrect thing like a wardrobe or a signal instead of a person. In that case, we count the picture as wrong.

For this phase, we took 160 pictures manually: 64 with people and 96 without people. In Figure 5, we show an example with three pictures: two with a person and another without people. Afterwards, we process these pictures with our Computer Vision module to detect the accuracy of our module using the three different models. With this test, we tried to evaluate the accuracy of each model that we use in our Computer Vision module.

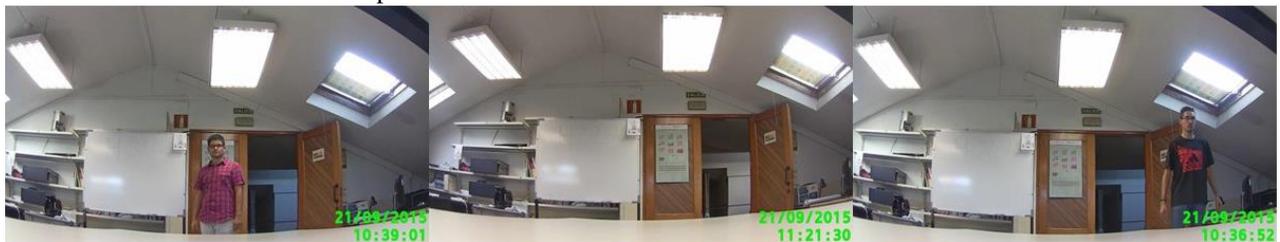
**Figure 5 Example with indoor pictures which took manually**

*2) Phase 2: Automatic pictures*

In the second phase, we programmed the camera to send a picture when the camera detected any movement in the area. In this case, the camera sent picture sequence from the first moment that the camera detects the movement until the last detection of the same movement. Then, we analysed all the sequence with each model to obtain if this method is valid to use the Computer Vision with an IP camera as a sensor. In Figure 6, we show some pictures of this sequences about how the camera detected the initial movement in the first picture and after, it continued sending one picture per second until the movement ceased. The camera is capable of sending from one picture per second to thirty pictures per second. However, to do the evaluation, we selected the maximum value in the camera configuration.

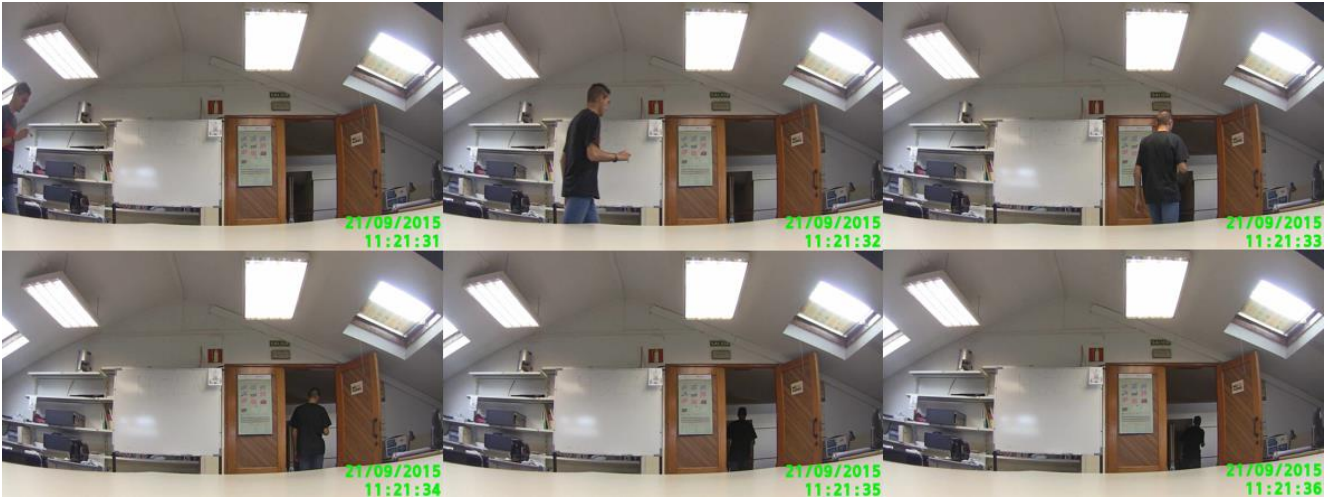
**Figure 6 Example of one of the picture sequences that the Canon IP camera send when detects movement**

The number of pictures depends on the movement time. The camera sent each picture to another web service created in Python with Flask to avoid the interaction of Midgar. The reason is that in this way we test only our Computer Vision module without a possible interference of the Midgar IoT platform. In this case, we intended to evaluate if we could obtain an improvement of the system using the camera sensor or maintain the same level. Besides, this way allows us a reduction of the network traffic, a reduction of the necessary process computer, and avoid using the streaming option. For this phase, we used 972 pictures divided into 17 sequences from which 8 sequences, which were composed by 817 pictures, were movements where appeared people and 9 sequences, which were composed by 155 pictures, were movements where did not appear people. Clarify, the sequences had many pictures without people because they have only a half body or an arm. Then, these sequences contained many invalidated pictures but these pictures were part of the sequence. For the negative sequences, we used the movement of different objects in front of the camera like balls, mugs, or papers.

*B. Results*

In this section, we are going to describe the results. In order to improve the understandability, we design tables and graphs that represent the results of testing the pictures with our Computer Vision module with each model and the results of the module classified in the different four groups:

- True Negative: pictures without people with a negative result. This is the best result for pictures without people because it means that the module does not detect people in pictures without people.
- False Positive: pictures without people with a positive result. This is the wrong case when we search people in pictures without people because this is when the module says that it found people.
- False Negative: pictures with people with a negative result. This is the worst result because is when the module analyses pictures with people but the module did not detect people in that picture.
- True Positive: pictures with people with a positive result. This is the best result when the module analyses pictures with people because this result appears when the module found people.

Next, in phase 1, we are going to describe the result with manual pictures. After that, we are going to show the second subsection, which contains the phase 2 with the result of using the camera sensor and analyse sequences of pictures instead of an isolated picture.

*1) Phase 1: Manual pictures*

Table 1 and Figure 7 show the result of the evaluation of the 160 manual pictures. These were divided into 96 pictures without people and 64 pictures with people, thus, the system can obtain a maximum of 96 true negatives or 96 false positives, and 64 true positives or 64 false negatives.

| Model | True Positive (64) | False Positive (96) | True Negative (96) | False Negative (64) |
| --- | --- | --- | --- | --- |
| Upper body | 7 / 10.94 % | 1 / 1.04% | 95 / 98.96% | 57 / 89.06% |
| Head and Shoulders | 22 / 34.38% | 0 / 0.00% | 96 / 100.00% | 42 / 65.63% |
| Frontal Face | 0 / 0.00% | 1 / 1.04% | 95 / 98.96% | 64 / 100.00% |

**Table 1 Results of the evaluation of the manual pictures**

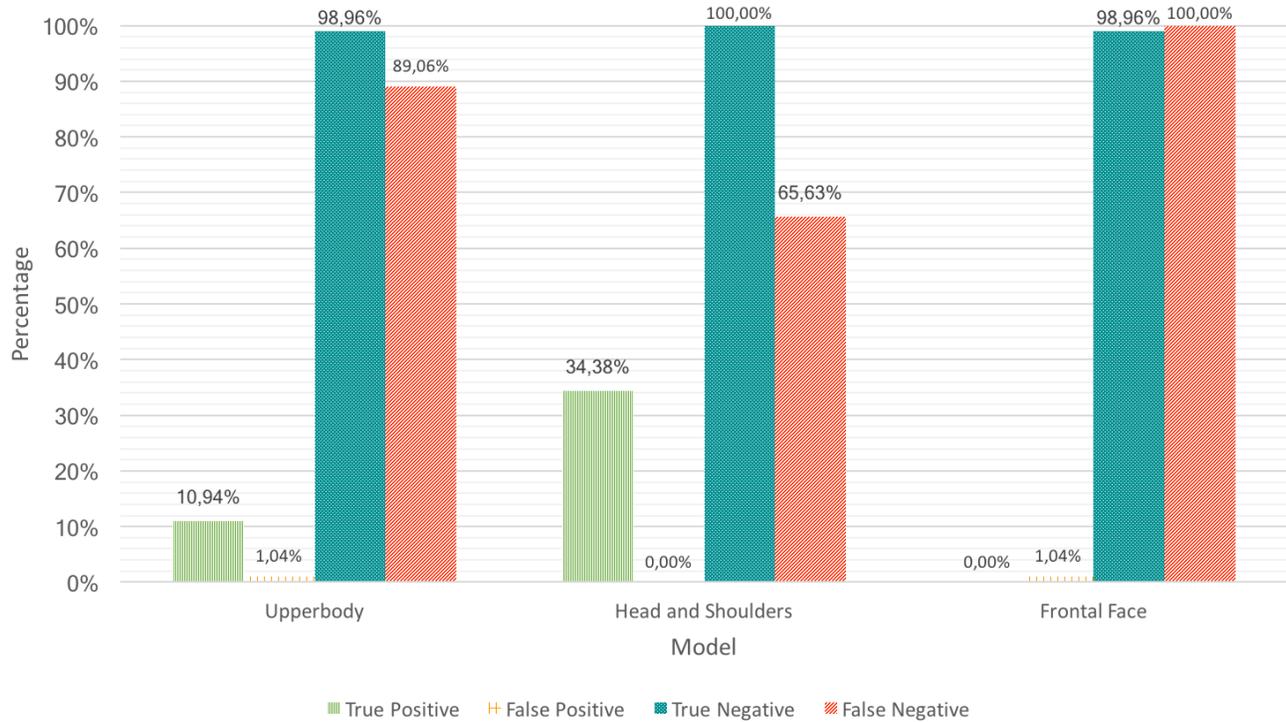

Figure 7 Results of the evaluation of the manual pictures

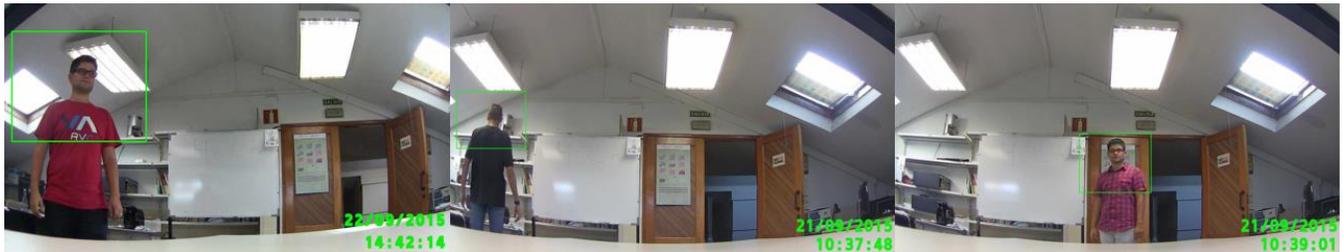

Figure 8 Three pictures of the true positive cases

By analysing Table 1 and Figure 7, we can suggest the following interpretations:
- After analysing the 96 pictures without people, the Computer Vision module interpreted 95 of these pictures as negative using the model for upper bodies detection. This represents a success of 98.96% of the total. Using the model for heads and shoulders detection, the Computer Vision module interpreted 96 pictures as negative which represents a success of 100.00%. Using the model for frontal faces detection, the Computer Vision module interpreted 95 pictures as negative that represents a success of 98.96%. The used model for detecting heads and shoulders obtained the best accuracy for the negative cases because had the highest success in the true negative cases whereas the model for detecting frontal faces obtained the worst accuracy for these cases.
- After analysing the 64 pictures with people, the Computer Vision module interpreted 7 of these pictures as positive using the model for upper bodies detection which represents a success of 10.94% of the total. In
- , we show three of the seven pictures that the Computer Vision module detected as true positive cases. Using the model for heads and shoulders detection, it interpreted 22 pictures as positive, which represents a success of 34,38%. Using the model for frontal faces detection, the Computer Vision module interpreted 0 pictures as positive which represents a success of 0.00%. The model for detecting frontal faces obtained the worst accuracy for the positive cases because the module did not detect people in pictures with people whereas the model for detecting heads and shoulders obtained the best accuracy for these cases although it is a very high unsuccessful result. Moreover, in some cases, the module interpreted as positive pictures with people but it detected other things instead of the people. These failures were interpreted as negative cases. We show two cases of this type in Figure 9.

- The best accuracy for detecting people in manual pictures was obtained by the model for detecting heads and shoulders because this model allowed to the Computer Vision module interprets as positive cases the highest number of positive pictures and the module interpreted as negative cases the highest number of negative pictures. Thus, the best model to use with manual pictures is the model for heads and shoulders. However, the accuracy for positive cases is too low to be able to demonstrate that pictures can be used as sensors.

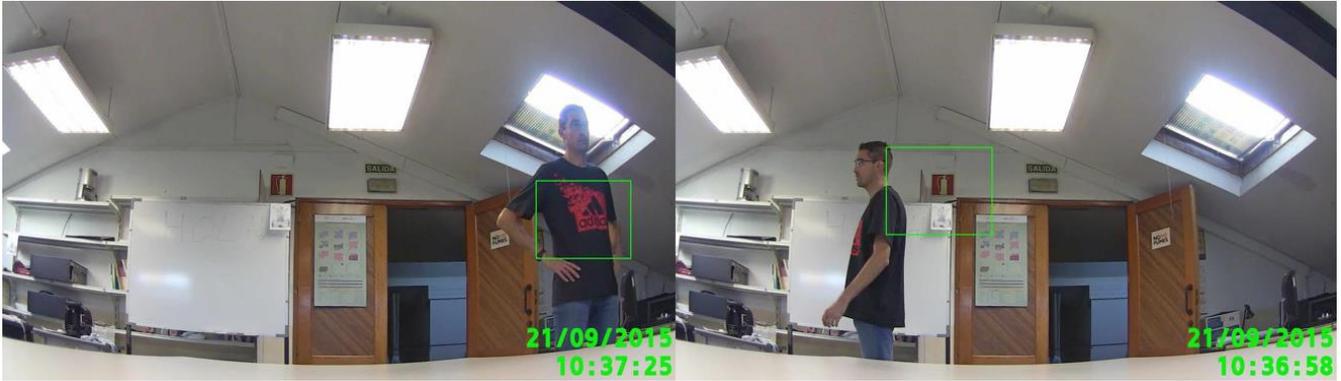

Figure 9 The two pictures that we discarded because the wrong detection

*2) Phase 2: Automatic pictures*

In this second phase, we analysed the pictures of the 17 sequences. We used 9 sequences without people and 8 with people. Thus, the system can obtain a maximum of 9 true negatives or 9 false positives, and 8 true positives or 8 false negatives. In this case, we analysed all the sequence as one item instead of analysing each picture separately. We can see some of the detected pictures in Figure 10.

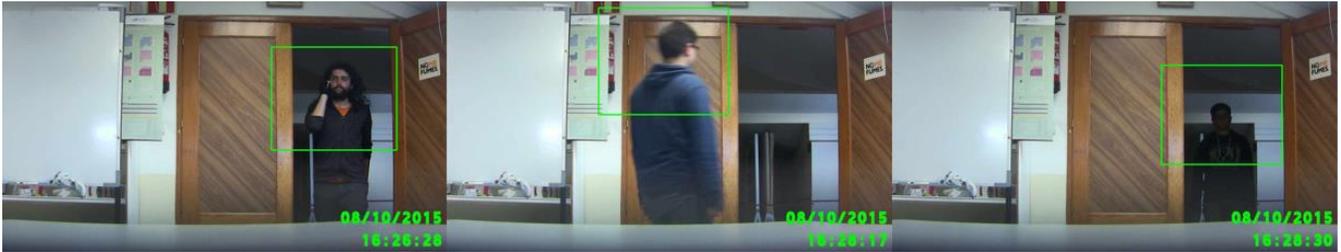

Figure 10 Collage with pictures of the True Positive sequences

In Table 2 and Figure 11, we show the results of applying our Computer Vision module to the picture sequences.

| Model | True Positive (8) | False Positive (9) | True Negative (9) | False Negative (8) |
|---|---|---|---|---|
| Upper body | 8 / 100.00% | 0 / 0.00% | 9 / 100.00% | 0 / 0.00% |
| Head and Shoulders | 7 / 87.50% | 0 / 0.00% | 9 / 100.00% | 1 / 12.50% |
| Frontal Face | 3 / 37.50% | 1 / 11.11% | 8 / 88.89% | 5 / 62.50% |

Table 2 Results of the Computer Vision module analysing the sequences

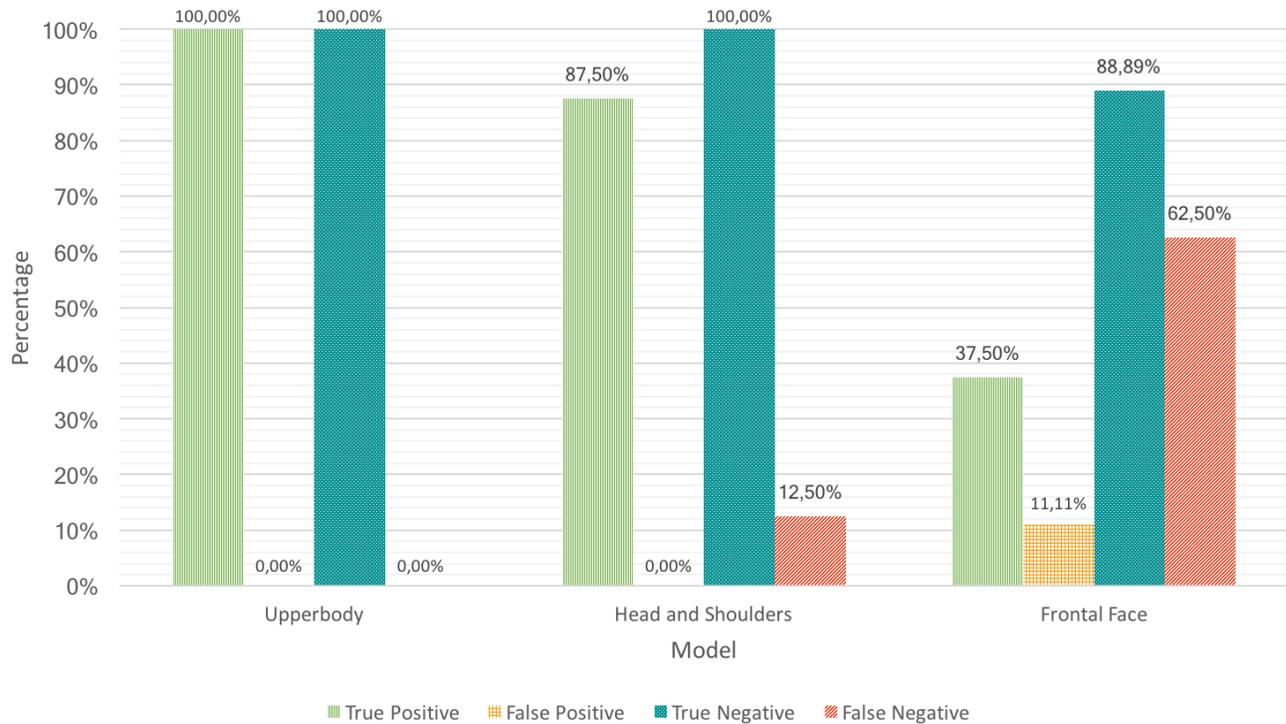

**Figure 11 Results of the Computer Vision module analysing the sequences**

We can suggest the next interpretations analysing the Table 2 and Figure 11:
- After analysing the 9 sequences without people, the Computer Vision module interpreted the 9 sequences as negative using the model for upper bodies detection which represents a success of 100.00%. Using the model for heads and shoulders detection, the module also interpreted the 9 sequences as negative therefore it also achieved a success of 100.00%. Using the model for frontal faces detection, the module interpreted 8 sequences as negative which represents a success of 88.89%. The models for detecting upper bodies and heads and shoulders obtained both the best possible accuracy for the negative cases whereas the model for detecting frontal faces obtained the worst accuracy for these cases although it is not a bad accuracy.
- After analysing the 8 sequences with people, the Computer Vision module interpreted the 8 sequences as positive using the model for upper bodies detection which represents a success of 100.00%. Using the model for heads and shoulders detection, it interpreted 7 sequences as positive which represents a success of 87.50%. Using the model for frontal faces detection, it interpreted 3 sequences as positive which represents a success of 37.50%. As it happened with manual pictures, the model for detecting frontal faces obtained the worst accuracy for the positive cases. However, the model that obtained the best accuracy is the model for detecting upper bodies.
- The Computer Vision module with the model for detecting upper bodies and our configuration obtained nine true positive cases for the nine positive sequences. It means the 100% of accuracy. This demonstrates that the use of picture sequences improves the accuracy in comparison with single pictures and the use of IP cameras as sensors is possible.

Table 3 shows the information about each sequence: name, if it contains people or not, the total pictures that the camera sent when detects the movement, which are the pictures that compose that sequence, the number and the percent of pictures that our module checked as picture with people, and the results according to our module for each model.

| Sequence Name | Sequence with People? | Total Pictures | Upper body | | | Head and Shoulders | | | Frontal face | | |
|---|---|---|---|---|---|---|---|---|---|---|---|
| | | | Identified Pictures | Detection % | Result | Identified Pictures | Detection % | Result | Identified Pictures | Detection % | Result |
| C1 | Yes | 78 | 5 | 6,41% | True Positive | 0 | 0,00% | False Negative | 0 | 0,00% | False Negative |
| C2 | Yes | 229 | 5 | 2,18% | True Positive | 9 | 3,93% | True Positive | 1 | 0,44% | True Positive |
| D1 | Yes | 53 | 17 | 32,08% | True Positive | 6 | 11,32% | True Positive | 0 | 0,00% | False Negative |
| D2 | Yes | 49 | 9 | 18,37% | True Positive | 8 | 16,33% | True Positive | 2 | 4,08% | True Positive |
| D3 | Yes | 54 | 19 | 35,19% | True Positive | 10 | 18,52% | True Positive | 0 | 0,00% | False Negative |
| D4 | Yes | 207 | 2 | 0,97% | True Positive | 4 | 1,93% | True Positive | 2 | 0,97% | True Positive |
| L1 | Yes | 84 | 13 | 15,48% | True Positive | 5 | 5,95% | True Positive | 0 | 0,00% | False Negative |
| L2 | Yes | 63 | 4 | 6,35% | True Positive | 4 | 6,35% | True Positive | 0 | 0,00% | False Negative |
| C | No | 12 | 0 | 0,00% | True Negative | 0 | 0,00% | True Negative | 0 | 0,00% | True Negative |
| F | No | 35 | 0 | 0,00% | True Negative | 0 | 0,00% | True Negative | 0 | 0,00% | True Negative |
| M | No | 9 | 0 | 0,00% | True Negative | 0 | 0,00% | True Negative | 0 | 0,00% | True Negative |
| P | No | 7 | 0 | 0,00% | True Negative | 0 | 0,00% | True Negative | 0 | 0,00% | True Negative |
| P1 | No | 7 | 0 | 0,00% | True Negative | 0 | 0,00% | True Negative | 0 | 0,00% | True Negative |
| P2 | No | 25 | 0 | 0,00% | True Negative | 0 | 0,00% | True Negative | 0 | 0,00% | True Negative |
| P3 | No | 17 | 0 | 0,00% | True Negative | 0 | 0,00% | True Negative | 0 | 0,00% | True Negative |
| SP | No | 30 | 0 | 0,00% | True Negative | 0 | 0,00% | True Negative | 1 | 3,33% | False Positive |
| T | No | 20 | 0 | 0,00% | False Positive | 0 | 0,00% | True Negative | 0 | 0,00% | True Negative |

**Table 3 Information about the sequences**

According to Table 3, we can suggest the next interpretation:
- The detection percent does not have relation with the number of pictures: 'C2' has 229 picture and it only obtained percentages below 5% of identified pictures. Meanwhile, 'D3', with 54 pictures, obtained a 35.19% with the model for detecting upper bodies. It depends on the pictures quality: people's position, picture sharpness, or full body shots because many pictures only contain the arm or a half body.

V. CONCLUSIONS

In this paper, we have presented a possible solution to use Computer Vision in the Internet of Things. This could allow using IP cameras and pictures as sensors. Besides, this could open the door to other similar applications like automating things using Optical Character Recognition (OCR), face detection, or gestures.

In this way, we have presented a possible architecture to integrate Computer Vision in an IoT platform, in our case Midgar. This Computer Vision module, analyses the picture, or, in our case, picture sequences, and returns to the IoT platform a Boolean result as if it was a button sensor. However, we have shown that if you have a weak model, the Computer Vision module could have a low accuracy. To deduce this, we tested our module using three different models for detecting upper bodies, heads and shoulders, and frontal faces. After evaluating our module with the three models, we obtained that the model for detecting head and shoulders is the best in the first phase, in which we analysed isolated pictures, with an accuracy of **100.00% for True Negatives** but a **34.38%** of accuracy for **True Positives**. Nevertheless, we proposed to analyse whole sequences of pictures, and we obtained much better results: the model for detecting upper bodies achieved an accuracy of **100% for True Positives** and **100.00% for True Negatives**. This was possible because we centred our module in the analysis of all the sequence of movement. Thus, the best model to analyse sequences is the model for detecting upper bodies. Moreover, we saw that the number of picture sequences does not affect the result because some cases with many pictures have the less detection percent.

In the case of **'C2' with 229 picture**, our module only obtained percentages **below 5% of identified pictures**. The contrary case was **'D3' with 54 pictures** because our module obtained a **35.19%** of accuracy with the model for detecting upper bodies. In this case, the important thing is the quality of the pictures. This is why, it is probably better to analyse sequences instead of isolated pictures, which is the thing that we do.

According to this article, we can say that it is possible the use of Computer Vision in the IoT. Besides, with a very interesting application. In addition, we have shown that we can use pictures as sensors using not a perfect or weak model because if we analyse all the sequence that we need, we can improve the results. However, it is better and more recommendable the use of a good model with the sequences because we could improve the accuracy.

Automating is one of the ways of the Internet of Things. In this way, if we could improve and add new functionalities with Computer Vision, we could facilitate the use of the Internet of Things in our daily life and we could create new ways to communicate us with our environment. In this case, we can use our research to automate and improve the security of our homes, towns, cities, industries, and the Earth.

## VI. FUTURE WORK

This proposal is a possible solution for some of the current problems of the IoT. There are many ways to continue this research and we are going to propose some of them in the next enumeration.:

- **Creating a Domain-Specific Language using Model-Driven Engineering to facilitate the Computer Vision**: In our proposal, we offered the possibility to detect only people in the pictures because you need to extract the features and train the model to detect other things. This is why offering a Domain-Specific Language to facilitate these steps in an IoT platform is one possibility.
- **Creating a system to control and define the different IP camera actions**: In this proposal, we used the IP camera as a sensor. In other proposals, the authors used the camera like an actuator, but only to take pictures. One possible way would be to develop an API to allow the total camera control like move it, change its properties, and so on.
- **IoT platforms scalability:** The Computer Vision needs many computer resources. Then, in the case that the computer received many pictures or has many IP cameras, the computer would need more time to process the whole information. This is why another future work way could be the study and test of different implementations in the IoT platforms to obtain a correct way of supporting the maximum number of IP cameras, or how to discard many repeated pictures.
- **Optical Character Recognition or Face detection:** Another field in Computer Vision is the character detection and the face detection. We could automate our doors to open it when it detects a determinate identification badge, license plate, or face. However, this is why we should check different secure options because the Computer Vision is a very problematic field due to the security that it needs.
- **Gesture recognition:** In this proposal, we analysed a sequence to detect people. Another possibility is to analyse the sequence to detect a determinate gesture to trigger an action. Then, we could use a picture sequence to detect a determinate gesture, which could throw a determinate action, previously assigned. It would allow using the pictures as sensors with different outputs.

## VII. ACKNOWLEDGEMENTS

This work was performed by the 'Ingeniería Dirigida por Modelos MDERG' research group at the University of Oviedo under Contract No. FC-15-GRUPIN14-084 of the research project 'Ingeniería Dirigida Por Modelos MDERG'. Project financed by PR Proyecto Plan Regional.

[52]     Etherios, Etherios, (2008).

[53]     ThingWorx<sup>TM</sup>, ThingWorx, (2015).

[54]     Carriots, Carriots, (2011).

[55]     Microsoft, Azure IoT Suit, (2015).

[56]     Amazon, AWS IoT, (n.d.).

[57]     IBM, IBM Internet of Things, (2015).

[58]     Paraimpu SRL, Paraimpu, (2012).

[59]     QuadraSpace, (2010).

[60]     A. Kansal, S. Nath, J. Liu, W.I. Grosky, SenseWeb : An Infrastructure for Shared Sensing, IEEE Multimed. 14 (2007) 8–13.

[61]     Microsoft, SenseWeb, Http://research.microsoft.com/en-Us/projects/senseweb/. (2008).

[62]     Department of Electrical and Electronic Engineering (University of Cagliari), SIoT, Http://platform.social-Iot.org/. (2012).

[63]     Oak Ridge National Laboratory, Sensorpedia, (2009).

[64]     B.L. Gorman, D.R. Resseguie, C. Tomkins-Tinch, Sensorpedia: Information sharing across incompatible sensor systems, 2009 Int. Symp. Collab. Technol. Syst. (2009) 448–454.

[65]     EVRYTHNG, EVRYTHNG, (2012).

[66]     Sen.se, Open.Sen.se, (2015).

[67]     IoBridge, Thingspeak, (2013).

[68]     Nimbits Inc., Nimbits, (2015).

[69]     CyberVision Inc., KAA, (2014).

[70]     W. Lorezo, R. Gonzalez-Crespo, A. Castillo-Sanz, A Prototype for linear features generalization, Int. J. Interact. Multimed. Artif. Intell. 1 (2010) 59.

[71]     D. Meana-Llorián, C. González García, B.C. Pelayo G-Bustelo, J.M. Cueva Lovelle, V.H. Medina García, IntelliSenses: Sintiendo Internet de las Cosas, in: 2016 11th Iber. Conf. Inf. Syst. Technol., AISTI, Gran Canaria, Spain, 2016: pp. 234–239.

[72]     Alejandro F. Reimondo, Haar Cascades, (n.d.).

[73]     M. Castrillón-Santana, O. Déniz-Suárez, L. Antón-Canalís, J. Navarro-Lorenzo, Face and Facial Feature Detection Evaluation: Performance Evaluation of Public Domain Haar Detectors for Face and Facial Feature Detection, Visapp 2008. (2008) 167–172.